\documentclass[runningheads]{llncs}
\usepackage[T1]{fontenc}

\usepackage[long]{optional} 
\usepackage{amsmath}
\usepackage{xcolor,graphicx}
\usepackage[pdftex]{hyperref}
\usepackage[capitalize,noabbrev]{cleveref}
\usepackage{array,booktabs,longtable}

\usepackage{color}

\begin{document}
\title{From Large Language Model Predicates to Logic Tensor Networks: Neurosymbolic Offer Validation in Regulated Procurement}
\titlerunning{Neurosymbolic Offer Validation in Regulated Procurement}
%
\author{
Cedric S. Haufe\inst{1,2}\orcidID{0009-0000-1172-7377} \and
Frieder Stolzenburg\inst{1,3}\orcidID{0000-0002-4037-2445}
}
\authorrunning{C. Haufe and F. Stolzenburg}

\institute{
Harz University of Applied Sciences, Wernigerode, Germany,
\email{\{cedric+haufe,fstolzenburg\}@hs-harz.de}
\and
Merseburg University of Applied Sciences, Merseburg, Germany,
\email{cedric.haufe@hs-merseburg.de}
\and
Senior Visiting Fellow, University of New South Wales (UNSW), School of Computer Science and Engineering, Sydney, Australia, \email{f.stolzenburg@unsw.edu.au}
}
\maketitle              
\begin{abstract}
We present a neurosymbolic approach\opt{long}{, i.e. combine symbolic and subsymbolic artificial intelligence,} to validating offer documents in regulated public institutions. We employ a language model to extract information and then aggregate it with an LTN (Logic Tensor Network) to make an auditable decision.
In regulated public institutions, decisions must be made in a manner that is both factually correct and legally verifiable.
Our neurosymbolic approach allows existing domain-specific knowledge to be linked to the semantic text understanding of language models. The decisions resulting from our pipeline can be justified by predicate values, rule truth values, and corresponding text passages. Our experiments on a real corpus show that the proposed pipeline achieves performance comparable to existing models, but its key advantage lies in its interpretability, modular predicate extraction, and explicit support for XAI (Explainable AI).

\end{abstract}

\section{Introduction}
In regulated areas such as public academic institutions and their internal procurement systems, simple document classification, e.g. whether a document is an offer or an invoice or something different, is not sufficient. Decisions on the extent to which a document can be considered a valid offer must be traceable and legally sound. \opt{long}{In the event of a dispute, it must be possible to trace the reasons that led to the rejection or acceptance \cite{Richmond.2023}.}
It must be clearly recognisable on the basis of which facts or characteristics a decision was made.
Conventional document classification, e.g. based on purely statistical or neural network models, does not provide any explanations directly linked to the original document~\cite{Rudin.2019}.
Classic explainable rule-based systems, on the other hand, require that the decision-relevant properties of a document are already available in a structured form. Accordingly, these must first be extracted from the text, which decouples them from the original document \cite{Besold.2022}. This leads to a gap between powerful but difficult-to-explain text models \cite{Rudin.2019} and precise but resource-intensive rule-based systems \cite{Garcez.2023}.

We bridge this gap with a neurosymbolic pipeline \cite{Besold.2022,Garcez.2023}
which combines predicate extraction based on a Large Language Model (LLM) with a Logic Tensor Network (LTN) for aggregation and final decision-making \cite{Badreddine.2022,Donadello.2017,Serafini.2016}. Unlike alternative neural-symbolic systems like DeepProbLog~\cite{manhaeve2021deepproblog} and NeurASP~\cite{yang2020neurasp} or unlike AI architectures such as LayoutLMv3~\cite{huang2022layoutlmv3} and Donut~\cite{kim2022donut}, our focus is evidence-preserving LLM predicate extraction for procurement documents.

In the first step, an LLM evaluates the truth values of predefined domain-specific predicates for each potential offer \opt{long}{\cite{Bodhwani.2025,Taubenfeld.2025}}\opt{short}{\cite{Bodhwani.2025}}.
An LTN aggregates these values together with coded assignment rules to make a decision about \texttt{IS\_VALID\_OFFER} and simultaneously provides predicate and rule truth values as an explanation.\opt{long}{ The rules map manual domain-specific experience and specifications of what makes an offer valid.} By storing the corresponding textual evidence in the predicate layer, it is thus possible to clearly identify which text passage has a high influence on the final decision. \opt{long}{Hence we obtain explanations in the sense of explainable AI (XAI). Our method does not serve as an end-to-end neurosymbolic training pipeline, but rather as a task-specific integration of various known modules, such as predicate-specific retrieval and LLM-based extraction, as well as an LTN-based fuzzy rule layer for the final decision. The individual modules are auditable, in some cases even with evidence chunks.
An overview of our proposed pipeline is shown in \cref{fig:pipeline} (see \cref{sec:architecture}).}

\section{Problem Definition and Data}

\subsection{Task Definition}

\opt{long}{In this article, we present a neurosymbolic approach for the automatic validation of offers in the context of regulated public academic institutions.
In the context of internal procurement, e.g. the purchase of a service or a specific product, one or more bids are submitted in the form of PDF files. An automatic decision must be made as to whether the individual documents submitted can be considered a \emph{valid offer}.}

Formally, let $\mathcal{D} = \{d_1, \dots, d_N\}$ be the set of potential offers to be considered, where each $d_i \in \mathcal{D}$ represents the text extracted from a PDF, including any Optical Character Recognition (OCR) if required. For each document $d_i$, there is a binary ground truth label
$y_i \in \{0,1\}$
where $y_i = 1$ means that $d_i$ can be accepted as a valid offer in the context described, and $y_i = 0$ means that it is not a valid offer (but rather, e.g. an invoice, delivery note, order confirmation, general price list, etc.) and therefore cannot be used for procurement.
\opt{long}{

}The primary task can thus be formulated as a binary classification task for documents: Given a document $d$, the corresponding label $y$ has to be predicted.
\opt{long}{In the context considered here, however, a pure binary label prediction is not sufficient in practice. The reasons that led to the decision must be traceable and verifiable in the event of a later dispute.}%
Accordingly, the decision must be based on explicit, domain-specific criteria and must be justifiable in the event of an audit or legal dispute \cite{Richmond.2023,Rudin.2019}. \opt{long}{It is therefore assumed that the prediction not only provides an assignment $d \mapsto y$, but can also be traced back to a set of semantically interpretable properties of the document (referred to as predicates) and an explicit decision logic about them. }We therefore model offer validation below as a binary classification task with additional requirements such as rule-based verifiability and explainability at the document level.

\subsection{Predicate Space and Semantic Structure}
\label{sec:predicates}

Through manual pre-review of original offers submitted in the context of procurement and internal procurement guidelines, we identify a set of $K$ domain-specific predicates that meaningfully characterise whether a document can be considered as an offer. In this context, $K = 8$ predicates were identified.
First, we model a vector of these domain-specific predicates for each document $d$:
\[
\mathbf{p}(d) = \big(p_1(d), \dots, p_K(d)\big)
\]
each of which describes a part of the central content-related properties of an offer. Each predicate $p_k(d)$ represents the degree to which a semantically clearly defined property is fulfilled and is later interpreted as a real truth value in $[0,1]$: The value $0$ corresponds to ‘does not apply’, values close to $1$ correspond to ‘clearly applies’.
We use the following predicates:
\opt{long}{
\begin{itemize}
\item \textbf{OFFER TITLE}: The document has a title or terms that clearly identify it as an offer.
\item \textbf{OFFER NUMBER}: There is an offer number, a reference or at least a similarly interpretable identifier.
\item \textbf{OFFER VALIDITY}: There is explicit information about the extent and duration of the potential offer's validity.
\item \textbf{RESERVATION CLAUSES}: Legal reservations or restrictions are formulated in the document.
\item \textbf{TERMS OF PAYMENT}: Terms of payment or general payment information are present, such as potential discounts, cash discounts, etc.
\item \textbf{TERMS OF DELIVERY}: Information is available on e.g. when a delivery can be made or to what extent preconditions must be met.
\item \textbf{SALES CONTACT}: The offer creator explicitly refers to a contact person who serves as the point of contact for this specific potential offer.
\item \textbf{NOT AN OFFER}: There is an indication that this is not an offer (but e.g. an invoice).
\end{itemize}
}
\opt{short}{

\noindent\begingroup
\small
\begin{list}{}{\setlength{\leftmargin}{3.75cm}
\setlength{\labelwidth}{3.55cm}
\setlength{\labelsep}{0.2cm}
\setlength{\itemsep}{0pt}
\setlength{\parsep}{0pt}
\setlength{\topsep}{2pt}}
\item[\texttt{OFFER\_TITLE}] Title or wording identifies the document as an offer.
\item[\texttt{OFFER\_NUMBER}] Offer number, reference or comparable identifier is present.
\item[\texttt{OFFER\_VALIDITY}] Validity period or validity scope is specified.
\item[\texttt{RESERVATION\_CLAUSES}] Legal reservations or restrictions are present.
\item[\texttt{PAYMENT\_TERMS}] Payment conditions are specified.
\item[\texttt{DELIVERY\_TERMS}] Delivery conditions or preconditions are specified.
\item[\texttt{SALES\_CONTACT}] A supplier or sales contact is given.
\item[\texttt{NOT\_OFFER}] Evidence indicates another document type, e.g. an invoice.
\end{list}
\endgroup

}
\opt{long}{
Conceptually, a truth value is determined for each predicate -- in our implementation, however, these are represented by derived channels (11 inputs in our case) in order to separate clear indications from vague/existing indications (see \cref{sec:ltn}).}

\subsection{Data Set and Annotation}
\label{sec:data}

Our data set is based on a corpus of documents, which originate from actual procurements by a German public university\opt{long}{ (Merseburg University of Applied Sciences)}.\opt{long}{ \cref{fig:example-offer} shows an anonymised real example of a valid offer document
from the corpus.}
All documents are available in PDF format and include both digitally created content and scanned documents.\opt{long}{ In terms of content, the documents cover a wide spectrum, from pure text documents to mixed documents with tables and diagrams.}
The average length of a document is approximately two pages. The corpus is linguistically heterogeneous, with a focus on German-language documents.
Each document $d$ was manually assigned a binary label $\texttt{IS\_VALID\_OFFER}(d) \in \{0,1\}$ which identifies the extent to which it is valid or invalid.
A valid offer is marked with $\texttt{IS\_VALID\_OFFER}$ = 1 and can be used in the real process.
Documents that clearly serve a different purpose\opt{long}{ (e.g. invoices, delivery notes, order confirmations, general price lists, internal forms or informal emails)} are marked with $\texttt{IS\_VALID\_OFFER} = 0$.

\opt{long}{\begin{figure}
\centering
\includegraphics[width=0.9\linewidth]{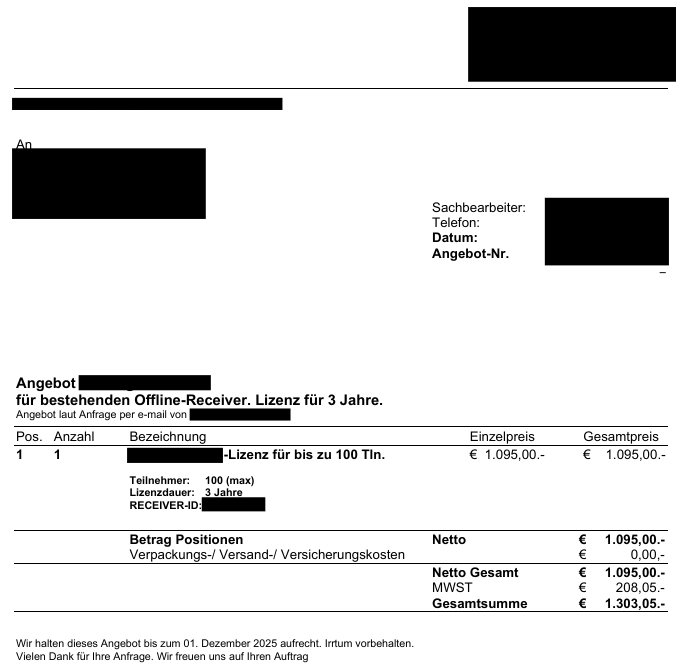}
\caption{Anonymised example of a valid offer document from the corpus.}
\label{fig:example-offer}
\end{figure}}

The annotation was carried out by the first author, partly in consultation with employees from internal specialist departments\opt{long}{ such as the budget department}, in order to correctly interpret and reflect internal guidelines. This is taken as ground truth.
The corpus considered here comprises a total of $N = 200$ documents, of which 35\,\% are annotated as valid offers\opt{long}{ (\texttt{IS\_VALID\_OFFER} = 1) and 65\,\% as non-offers (\texttt{IS\_VALID\_OFFER} = 0)}.

\opt{long}{We use repeated stratified 5-fold cross-validation with 5 repetitions \cite{Kohavi.1995}: The 200 documents are divided into 25 different stratified training/test parts, with the proportion of valid offers corresponding to the overall distribution.
In each fold, 80\,\% of the documents serve as the training set and 20\,\% as the test set. The evaluation metrics (given in \cref{sec:metrics}) are computed as averages over the 25 folds. The quantitative evaluation is based on the primary task (binary classification) and the binary decision \texttt{IS\_VALID\_OFFER}.}

The documents contain personal data as defined by European data protection law\opt{long}{ (e.g. names and contact details of university employees and suppliers, signatures, addresses, telephone numbers, etc.)}.\opt{long}{ In accordance with the General Data Protection Regulation (GDPR) and the related internal guidelines of the academic institution, these documents may only be processed within the university's IT infrastructure and may not be passed on to external service providers. Therefore, all calculations discussed in this article are performed using locally deployed LLMs. At no point content is sent to external LLM APIs.}

\section{Method: The Neurosymbolic Pipeline}
\label{sec:method}

\subsection{Architecture Overview}
\label{sec:architecture}

The potential offers submitted in the course of a procurement process are first prepared. For this purpose, the text contained therein is extracted or, if necessary, OCR is performed. The locally deployed LLM is then used as a predicate layer to determine truth values for the domain-specific predicates (see \cref{sec:predicates}).\opt{long}{ \cref{fig:pipeline} provides an overview of the overall architecture including the LTN which is used for the final decision.}
\opt{long}{
\begin{figure}[ht]
\centering
\includegraphics[width=0.8\linewidth]{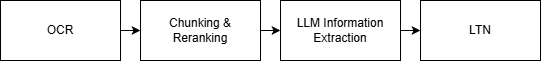}
\caption{Overview of the pipeline: Incoming potential offers are segmented, evaluated by an LLM, and then used by an LTN to make the final decision.}
\label{fig:pipeline}
\end{figure}}

Since these potential offers usually contain several pages and numerous layout elements\opt{long}{ (text, tables, diagrams)}, the LLM does not work directly with the entire raw text. Instead, this raw text is divided into a series of meaningful text excerpts\opt{long}{ (chunks)}.\opt{long}{ For this purpose, it is divided into overlapping text areas.}

For each predicate $p_k$, we then formulate a predicate-specific query and evaluate the chunks based on this.
We use a multi-stage search module for this. First, a lexical ranking of the chunks is determined based on the Best Matching ranking algorithm (BM25) \cite{Robertson.2009}. This is followed by a semantic re-ranking, namely embedding similarity and an LLM-based cross-encoder.
Only the chunks with the best evaluation are used to determine the truth values of the respective predicate \cite{Lewis.2020,Lin.2022}.\opt{long}{

For the predicate layer, we compare two different LLM-based extraction methods that follow different self-evaluation and aggregation strategies: a multi-class self-reflection approach (MCSR) \cite{Bodhwani.2025}\opt{long}{ and a confidence-informed self-consisten\-cy approach (CISC) \cite
{Taubenfeld.2025}}.
Both provide a numerical value in $[0,1]$ for each predicate within a potential offer, which is interpreted as a soft truth value. However, they differ in terms of the type of LLM queries and aggregation.}
After that, an LTN is used \cite{Badreddine.2022,Donadello.2017,Serafini.2016} to make a decision about the validity of the potential offers based on these truth values.\opt{long}{ LTNs integrate fuzzy logic rules with continuous predicate values in a differentiable framework.

In our LLM+LTN variants (see \cref{sec:pipeline_mcsr}), we train gating parameters with a supervised Binary Cross-Entropy (BCE) loss function.} The predicate values $\mathbf{p}(d)$ are used as continuous inputs to apply explicit domain-specific rules. \opt{long}{These can formally describe the predicates to form a global decision \texttt{IS\_VALID\_OFFER} (e.g. that a valid offer usually has a title, an offer number, a validity period, payment and delivery terms, and at the same time there are no strong indicators for \texttt{NOT\_OFFER}).
To evaluate the performance of the approach and the contribution of the predicate layer, the LTN, and the logical backend, we compare the presented neurosymbolic pipeline in different variations as well as with current alternatives (see \cref{sec:models}).}

\subsection{LLM-Based Predicate Extraction}
\label{sec:llm-predicates}

The predicate layer determines a numerical value $p_k(d) \in [0,1]$ for each potential offer $d$ and each predicate $p_k$.\opt{long}{ This expresses the extent to which the corresponding property of the predicate is fulfilled or applies.}
The calculation is performed through structured interaction with an LLM which uses selected chunks to make judgements and express confidence levels for each predicate.
In all experiments, we primarily use a 14B instruction model from Qwen2.5 (\texttt{Qwen2.5-14B-Instruct}) \cite{Bai.2023} as the LLM within the predicate layer. \opt{long}{The model is used exclusively via prompting. No further fine-tuning is performed.} For each predicate, we perform up to three calls to the 14B model to obtain syntactically valid and contextually usable JSON output.\opt{long}{ If no valid JSON is generated in these three attempts, a one-time fallback call is made to a larger 32B LLM (\texttt{Qwen2.5-32B-Instruct}). This happened in 1.56\% of cases. The resulting outputs provide truth values, which are then used as input for the LTN decision layer.}\opt{short}{ If this fails, a fallback call is made to a larger 32B LLM (\texttt{Qwen2.5-32B-Instruct}). }

\opt{long}{
\subsubsection{MCSR-Oriented Rating Estimation.}
\label{sec:pipeline_mcsr}}

\opt{short}{Our pipeline }\opt{long}{The first variant} is based on the Multi-Class Self-Reflection approach (MCSR)~\cite{Bodhwani.2025}.\opt{long}{ Here, structured self-reflection is used to determine more reliable confidence estimates.} For each predicate $p_k$, we use a three-level ordinal scale with the classes $c \in \{0,1,2\}$ (where 0 = not fulfilled, 1 = partially fulfilled or uncertain, 2 = clearly fulfilled or present). For a potential offer $d$ and a predicate, the model receives a predicate-specific query describing the classes and their semantic meaning (e.g. ‘no offer number recognisable’, ‘some clues are present but unclear’, ‘clearly marked as offer number’).
In a single query (‘evaluate – reflect – conclude’), the LLM first evaluates each class separately and outputs an initial confidence estimate together with evidence. This evaluation is then reflected upon and the class-related confidence values \opt{long}{(\textit{reflected confidences}) }are revised on this basis. Finally, the class with the highest confidence is selected as the winning class~\cite{Bodhwani.2025}.

\opt{long}{In the \textit{BestConf} variant, only the winning class is considered: the}\opt{short}{The} reflected confidence of this class is forwarded as a truth value to the decisive LTN layer.\opt{long}{ All non-winning classes are treated as 0.
In contrast, in the \textit{TopProb} variant, the probability distribution of all classes is first normalised. The resulting predicate value $p_k(d)$ is then derived from the normalised probability mass of the winning class. Therefore, TopProb assigns lower values because the model distributes the probability mass across several classes.} The resulting values $p_k(d) \in [0,1]$ are used as fuzzy truth values in the LTN decision layer.

\opt{long}{\subsubsection{CISC-Oriented Predicate Estimation.}

The second variant follows the confi\-dence-based self-consistency principle (CISC) \cite{Taubenfeld.2025}, which generates multiple response paths for inference tasks and evaluates them with confidence levels. In our context, we use CISC in a predicate- and chunk-based manner: for each potential offer $d$, the text is divided into overlapping chunks \cite{Lewis.2020}, and for a given predicate $p_k$, we use predicate-specific retrieval to select those chunks that are likely to contain relevant evidence.
The LLM now evaluates each of these chunks binarily (valid/invalid) to determine the extent to which the corresponding predicate applies/occurs. At the same time, this decision is underpinned by LLM self-assessment with a numerical confidence score.
This process is carried out across the chunks and in multiple iterations. This results in a series of binary individual decisions with corresponding confidence scores.
The truth value of the predicate $p_k(d)$ is then determined from a weighted aggregation of these individual predicate-specific evaluations. In contrast to the MCSR variant, which explicitly models the ordinal classes, the CISC variant works with fine-grained, confidence-weighted binary pieces of evidence that are combined across multiple blocks.}

\bigskip\noindent All prompts and implementation details are available as repository at \href{https://github.com/cedricshaufe/From-Large-Language-Model-Predicates-to-Logic-Tensor-Networks}{GitHub}\opt{short}{ (see also \cite{HS26a})}.

\subsection{LTN-Based Decision Logic}
\label{sec:ltn}

The predicate layer described in \cref{sec:llm-predicates} assigns a vector with continuous truth values to each potential offer $d$:
\[
\mathbf{p}(d) \in [0,1]^8
\]
However, in our implementation, these eight predicates are mapped by 11 derived input channels\opt{long}{ (see \cref{tab:channels})}.\opt{long}{ These channels can capture finer gradations and thus contain more information.} This allows clear evidence to be distinguished from vague/existing evidence for selected predicates, while still allowing them to be considered together.
\opt{long}{In the following, we use the shorthand notation
\[
T, N, V, R, P, D, S, \textit{NOT}
\]
for the predicates \texttt{OFFER\_TITLE}, \texttt{OFFER\_NUMBER},
\texttt{OFFER\_VALIDITY}, \texttt{RESERVATION\_ CLAUSES},
\texttt{PAYMENT\_TERMS}, \texttt{DELIVERY\_TERMS},
\texttt{SALES\_CONTACT} and \texttt{NOT\_OFFER}. Here,
for example, $T(d)$ denotes the degree of truth estimated for the predicate \texttt{OFFER\_TITLE} in the potential offer $d$.
The statement \texttt{IS\_VALID\_OFFER} is mapped by a target predicate $O$. Its truth value $o(d) \in [0,1]$ describes the extent to which an offer is valid.}

To aggregate these predicates, we use an LTN in the sense of Real Logic
\cite{Badreddine.2022,Donadello.2017,Serafini.2016}. In Real Logic, predicates
and formulas are assigned truth values in $[0,1]$, and fuzzy logic operators are
defined via differentiable T-norms, e.g. Gödel, Łukasiewicz, and Product
\cite{Hajek.1998}.%
\opt{long}{\[
\begin{aligned}
\textbf{Gödel:}\quad
& a\wedge b:=\min(a,b)\\
& a\vee b:=\max(a,b)\\
& a\rightarrow b:=
\begin{cases}
1, & a\le b,\\
b, & a>b
\end{cases}
\\[2pt]
\textbf{Product:}\quad
& a\wedge b:=a\,b\\
& a\vee b:=a+b-a\,b\\
& a\rightarrow b:=\min\!\left(1,\frac{b}{a}\right)
\\[2pt]
\textbf{\L ukasiewicz:}\quad
& a\wedge b:=\max(0,a+b-1)\\
& a\vee b:=\min(1,a+b)\\
& a\rightarrow b:=\min(1,1-a+b)
\end{aligned}
\label{eq:fuzzy-ops}
\]
A formula is considered to be more ‘true’ the better the corresponding
combination of predicate values matches the intended logical structure
\cite{Badreddine.2022,Serafini.2016}.
\begin{table}[h]
  \centering
  \small
\caption{The 11 channels derived from the 8 predicates. These 11 channels are used in the LTN decision layer described.}
\label{tab:channels}
\begin{tabular}{ll}
\toprule
\textbf{Symbol} & \textbf{Channel} \\
\midrule
$T_c$   & Title\_clear            \\
$N_c$   & Number\_clear            \\
$N_p$   & Number\_present         \\
$V_c$   & Validity\_clear          \\
$V_v$   & Validity\_vague         \\
$R_c$   & Reservation\_clear      \\
$P_p$   & Payment\_present         \\
$D_p$   & Delivery\_present        \\
$S_p$   & SalesContact\_present    \\
$NOT_s$ & NotOffer\_strong       \\
$NOT_v$ & NotOffer\_vague       \\
\bottomrule
\end{tabular}
\end{table}
\subsubsection{Basic aggregation of proofs.}
In the implementation, these are represented by a fixed derived input vector $\tilde{\mathbf{p}}(d)\in[0,1]^{11}$, which separates \emph{clear} vs. \emph{vague/existing} evidence for selected predicates (e.g. validity and offer number).} Let $\wedge$, $\vee$ and $\neg$ be the fuzzy conjunction, disjunction and
negation induced by the selected logic backend\opt{long}{ (Gödel, Product or Łukasiewicz)}.
In our pipeline, the influence of selected channels is mapped by learnable
\emph{smooth gates} $g_i=\sigma(\alpha_i)\in(0,1)$ where $\sigma$ denotes the
sigmoid function and $\alpha_i$ is a hyperparameter, i.e. $\tilde
p_i(d)=g_i\,p_i(d)$.
\opt{long}{Here, $p_i(d)$ denotes the $i$-th component of the derived input vector before gating and $\tilde p_i(d)$ denotes the forwarded gating value.}
\opt{long}{\textit{NOT}}\opt{short}{\textit{NOT\_OFFER}} is mapped as negative evidence by a separate learnable gate.

Using the fuzzy operators of the selected backend, we first determine a core for the positive evidence
$\mathrm{PosCore}(d)$ and then a core for the negative evidence $\mathrm{NegCore}(d)$
(\opt{long}{\textit{NOT}}\opt{short}{\textit{NOT\_OFFER}}). The base offer value is defined as
\[
O_{\text{base}}(d) \;=\; \mathrm{PosCore}(d)\ \wedge\ \neg\,\mathrm{NegCore}(d),
\]
which yields $O_{\text{base}}(d)\in[0,1]$.
\opt{long}{In our implementation, we form $\mathrm{PosCore}(d)$ from a disjunction of
central domain-specific typical offer indicators.
$\mathrm{NegCore}(d)$ corresponds to (\textit{NOT}) as counterevidence.

\subsubsection{Explicit offer rules.}}
A series of fuzzy logic implications are formulated using predicates. This means that the final decision can also be supported by domain-specific rules. These rules describe when an offer is valid in this context. \opt{long}{This applies when several positive characteristics are present at the same time and there is no strong evidence to the contrary (see \cref{tab:rules}).}
\opt{long}{\begin{table}[b]
  \centering
  \caption{The following rules are formulated from the 11 derived channels.}
  \label{tab:rules}
\vspace*{-5ex}
\begin{equation}
\mathrm{PosFeature}(d)
=
T_c(d)\vee N_p(d)\vee V_v(d)\vee R_c(d)\vee
P_p(d)\vee D_p(d)\vee S_p(d).
\end{equation}
\begin{align}
R_1(d):\;& T_c(d)\rightarrow O_{\mathrm{base}}(d),\\
R_2(d):\;& \bigl(V_c(d)\wedge P_p(d)\bigr)\rightarrow O_{\mathrm{base}}(d),\\
R_3(d):\;& \bigl((T_c(d)\vee N_c(d))\wedge \neg NOT_s(d)\bigr)\rightarrow O_{\mathrm{base}}(d),\\
R_4(d):\;& \mathrm{PosFeature}(d)\rightarrow O_{\mathrm{base}}(d),\\
R_5(d):\;& NOT_s(d)\rightarrow \neg O_{\mathrm{base}}(d),\\
R_6(d):\;& NOT_v(d)\rightarrow \neg O_{\mathrm{base}}(d).
\end{align}
\vspace*{-5ex}
\end{table}}These implications are evaluated using a fuzzy implication function\opt{long}{, e.g. based on Łukasiewicz logic} \cite {Hajek.1998}. A truth value $R_k(d) \in [0,1]$ is assigned to each rule.
 \opt{long}{This indicates the extent to which the rule is taken into account in the current decision \cite{Badreddine.2022,Serafini.2016}.
 The truth values of the rules, the predicate values or, more broadly, the predicate proofs can be tracked in a practical environment in order to make the corresponding decision legally comprehensible. These rule truth values can later be output together with the predicate values as part of the explanation of a decision, analogous to the use of LTN rules in semantic image interpretation \cite{Donadello.2017}.

\subsubsection{Final Decision and Training.}
The LTN decision layer outputs a base evaluation $\mathrm{O}_{\text{base}}(d)$. This is calculated from the gated predicate channels using the selected fuzzy logic backend.}%
\opt{short}{The LTN decision layer outputs a base evaluation $\mathrm{O}_{\text{base}}(d)$, calculated from the gated predicate channels using the selected fuzzy logic backend.}
\opt{long}{In our implementation, this base score is the decision on how valid an offer is. }\texttt{IS\_VALID\_OFFER} is decided by applying a threshold to the base score. The threshold to be used is determined on the training data of each fold by optimising the $F_1$-score of the positive class.
The learnable gating parameters are trained in a supervised manner by minimising the \opt{short}{binary cross-entropy} (BCE) on the training split of each fold.
\opt{long}{Furthermore, rule truth values are calculated from the fuzzy implications and reported together with predicate values to support verifiability and explainability. We are investigating a modular neurosymbolic approach in which predicate values extracted by an LLM are mapped to an LTN-based fuzzy rule layer, rather than training the entire pipeline end-to-end. This is intended to provide transparent decision support.}

\section{Experimental Setup}
\label{sec:experiments}
\opt{long}{
\subsection{Research Questions}

Our experiments and calculations aim to quantify the advantages of the proposed neurosymbolic pipeline over alternatives and to demonstrate a way to enable explainable decisions in the current context using neurosymbolic AI. Our focus is on the following research questions:

\begin{itemize}
\item \textbf{RQ1:} What trade-off is involved in replacing directly learned predicates with evidence-based LLM-derived predicate estimates?
\item \textbf{RQ2:} How does classification performance compare when using an MCSR-oriented pipeline versus a CISC-oriented one?
\item \textbf{RQ3:} What influence does the choice of logical backend have on decision quality when using the same classification method?
\end{itemize}
}

\subsection{Model Variants}
\label{sec:models}
\opt{long}{
To answer the research questions, the following variants are implemented:

\begin{itemize}
\item \textbf{MCSR+LTN:} A pipeline that uses an MCSR-oriented approach with ordinal predicate estimation
(\cref{sec:llm-predicates}) together with LTN-based decision logic
(\cref{sec:ltn}).
\opt{long}{
\item \textbf{CISC+LTN:} A pipeline that uses a CISC-oriented approach with confidence-weighted predicate estimates. An LTN is used to aggregate the truth values.
}
\item \textbf{LTN}: Classic LTN configuration without an explicit LLM predicate layer. The truth values of the predicates are learned directly from the extracted text (e.g. TF-IDF vectors). We use an analogue set of rules to check the validity of the offer.

\item \textbf{BERT:} Serves as a purely neural baseline, where a finely tuned German-language
BERT model (\texttt{bert-base-german-cased} \cite{Devlin.2019}) predicts \texttt{IS\_VALID\_ OFFER}
directly from the extracted text.

\item \textbf{LLM (BM25+Semantic-CE):} A single LLM is used to determine the extent to which something constitutes an offer, based on the existing predicates that serve as criteria.
\opt{long}{
\item \textbf{Information Extraction (IE) + deterministic Rules:} Pattern recognition is used to generate predicate scores, which are then aggregated according to a formula to produce an overall score that determines whether an offer is valid.
}
\end{itemize}
}
\opt{short}{
We evaluate \textbf{MCSR+LTN} as a combination of MCSR-based predicate estimation and LTN-based decision logic, \textbf{LTN} as a classic baseline, \textbf{BERT} as a purely neural baseline and \textbf{LLM (BM25+Semantic-CE)} as a direct LLM baseline.
}For the LTN-based variants, we examine several configurations of the logical
backend (e.g. {\L}ukasiewicz logic and product-based T-norms) to assess their influence \cite{Badreddine.2022,Hajek.1998,Serafini.2016}.

\subsection{Evaluation Report}
\label{sec:eval-protocol}

Since the available data set is small and slightly unbalanced (\opt{longt}{35\,\% valid offers,
}see \cref{sec:data}), we use stratified 5-fold cross-validation.  This involves performing 5 repetitions, resulting in 25 folds\opt{ short}{with a similar class distribution}.%
\opt{long}{ All 25 folds have a similar class distribution from the 200 potential offers.}
In each fold, four folds serve as the training data set and one fold serves as the
test data set. The metrics obtained are averaged across the 25 test folds.\opt{long}{ This cross-validation is an established method for estimating generalisation performance and model selection. Stratified variants reduce
the variance and class shift between folds \cite{Kohavi.1995}.}
The hyperparameters of the models\opt{long}{ (e.g. thresholds or predicate weights in the LTN)} are determined exclusively on the basis of the training data of one fold. The test data of one fold is only used for the final evaluation. \opt{short}{Since \texttt{IS\_VALID\_OFFER} is a binary predicate with unbalanced class distribution
(35\,\% positive class), we primarily specify the F$_1$-score of the positive class
$\mathrm{F1}_{\text{pos}}$.}

\opt{long}{
\subsection{Evaluation Metrics}
\label{sec:metrics}

Since \texttt{IS\_VALID\_OFFER} is a binary predicate with unbalanced class distribution
(35\,\% positive class), we primarily specify the F$_1$-score of the positive class
$\mathrm{F1}_{\text{pos}}$, i.e. the harmonic mean of precision and recall
\cite{Powers.2011,Sokolova.2009}. Furthermore, the accuracy, precision and recall of the
positive class are specified. All metrics are calculated per test fold across the 25 folds \cite{Kohavi.1995}.}

\section{Empirical Results}
\label{sec:results}
\opt{long}{

\subsection{Evaluation}
We use the F$_1$ score of the positive class
(\emph{valid offers}) as the main metric, supplemented by accuracy, precision and recall of the
positive class. Since the class distribution is unbalanced, we report the F$_1$ score of the
positive class as the primary metric \cite{Powers.2011}.
Since only 35\,\% of the examples are positive, a pure accuracy consideration would be
highly distorted.

For each pipeline, we consider (if possible) three variants of fuzzy logic
(Gödel, Product, {\L}ukasiewicz) with a threshold, which is adjusted based on the F$_1$ score of the positive class on the respective
training data.
In \cref{tab:main-results}, we report the
configuration with the best mean F$_1$ score on the test folds for each pipeline.

For the LLM-based predicate variants (MCSR, CISC), LLM-based predicate estimation with the configuration described in \cref{sec:llm-predicates} is primarily implemented using a 14B instruction model (\texttt{Qwen2.5-14B-Instruct}).}

\subsection{Overall Comparison of Pipelines}

\cref{tab:main-results} shows a post-hoc summary of the best F$_1$ results for each pipeline.
The pure LTN variant achieves an average
F$_1$ value of $0.899$ for the positive class and accordingly serves as the neurosymbolic baseline.
\opt{long}{The combination of CISC predicate extraction and LTN logic achieves a F$_1$ value of $0.782$, but still performs worse than the LTN baseline.}

\begin{table}[ht]
  \centering
  \caption{Post-hoc average test performance of the approaches considered over
    25 test folds (5 stratified splits $\times$ 5-fold cross-validation,
    $N = 200$, 35\,\% valid offers).}
  \label{tab:main-results}
  \begin{tabular*}{\linewidth}{@{\extracolsep{\fill}} l l c @{}}
    \toprule
    Pipeline & Fuzzy-Logic & F$_1$ (pos.) \\
    \midrule
    LTN            & {\L}ukasiewicz   & $0.899 \pm 0.064$ \\
 \opt{long}{   CISC           & {\L}ukasiewicz    & $0.782 \pm 0.060$ \\}
    MCSR\opt{long}{-BestConf}\opt{short}{+LTN}  & Product          & $0.874 \pm 0.043$ \\
\opt{long}{    MCSR-TopProb   & {\L}ukasiewicz   & $0.849 \pm 0.049$ \\ }
    BERT           & --               & $0.859 \pm 0.097$ \\
        LLM (BM25+Semantic-CE)\opt{long}{ - Qwen2.5:14b-instruct}          & --               & $0.843 \pm 0.059$ \\
   \opt{long}{  LLM (BM25+Semantic-CE) - Qwen2.5:32b           & --               & $0.884 \pm 0.058$ \\}
  \opt{long}{  IE + deterministic Rules           & --               & $0.807 \pm 0.056$ \\}

    \bottomrule
  \end{tabular*}
\end{table}
MCSR-based predicate recognition does not surpass the classic-LTN configuration on this corpus\opt{long}{.
MCSR-BestConf+LTN}\opt{short}{ and} achieves $0.874$\opt{long}{, while the
MCSR-TopProb+LTN pipeline achieves $0.849$}.
\opt{long}{The standard deviations across the 25 folds remain moderate
(approx. $0.04$ to $0.05$).}
In comparison, the sole use of the same language model and retrieval with an F$_1$ value of $0.843$ performs slightly worse than the MCSR approach\opt{long}{ but better than the CISC pipeline}.\opt{long}{ Only the use of a larger LLM (\texttt{Qwen2.5:32B}) results in an increase in F$_1$ to $0.884$, thereby outperforming the MCSR approach.
The use of pattern recognition with the subsequent application of deterministic rules results in an F$_1$ value of $0.807 \pm 0.056$ and thus performs worse than the MCSR approach described above.}
As a purely neural reference, we additionally use a BERT-based document classification (\text{bert-base-german-cased} \cite{Devlin.2019}), which predicts \texttt{IS\_VALID\_OFFER} directly from the full texts.
With an identical cross-validation protocol, BERT achieves a mean F$_1$ value of $0.859 \pm 0.097$ (\cref{tab:main-results}).\opt{long}{ The complete metrics (precision, recall, F$_1$, and accuracy) for all
pipelines are listed in \cref{tab:detail-results}.

\begin{table}[ht]
  \centering
  \small
  \caption{Average test performance of the approaches considered over
    25 test folds (5 stratified splits $\times$ 5-fold cross-validation,
    $N = 200$, 35\,\% valid offers). The
    F$_1$ score of the positive class is given (mean $\pm$ standard deviation).}
  \label{tab:detail-results}
  \begin{tabular*}{\linewidth}{@{\extracolsep{\fill}} l c c c c @{}}
    \toprule
    Fuzzy-Logic & Precision & Recall & F$_1$ & Accuracy \\
    \midrule
    \multicolumn{5}{l}{\textbf{LTN}} \\
    \midrule
    Gödel          & 0.837 & 0.960 & $0.892 \pm 0.055$ & 0.916 \\
    {\L}ukasiewicz & 0.862 & 0.946 & $\boldsymbol{0.899 \pm 0.064}$ & 0.924 \\
    Product        & 0.863 & 0.943 & $0.898 \pm 0.058$ & 0.923 \\
    \midrule
    \multicolumn{5}{l}{\textbf{MCSR-BestConf}} \\
    \midrule
    Gödel          & 0.768 & 0.908 & $0.819 \pm 0.064$ & 0.857 \\
    {\L}ukasiewicz & 0.781 & 0.896 & $0.827 \pm 0.055$ & 0.865 \\
    Product        & 0.832 & 0.929 & $\boldsymbol{0.874 \pm 0.043}$ & 0.903 \\
    \midrule
    \multicolumn{5}{l}{\textbf{MCSR-TopProb}} \\
    \midrule
    Gödel          & 0.834 & 0.830 & $0.825\pm 0.051$ & 0.875 \\
    {\L}ukasiewicz & 0.839 & 0.874 & $\boldsymbol{0.849\pm 0.049}$ & 0.889 \\
    Product        & 0.778 & 0.916 & ${0.836\pm 0.054}$ & 0.871 \\
    \midrule
    \multicolumn{5}{l}{\textbf{CISC}} \\
    \midrule
    Gödel          & 0.748 & 0.807 & ${0.770\pm 0.062}$ & 0.824 \\
    {\L}ukasiewicz & 0.784 & 0.796 & $\boldsymbol{0.782\pm 0.060}$ & 0.840 \\
    Product        & 0.774 & 0.798 & ${0.778\pm 0.054}$ & 0.836 \\
    \midrule
    \multicolumn{5}{l}{\textbf{BERT-baseline}} \\
    \midrule
    --             & 0.872 & 0.865 & $\boldsymbol{0.859 \pm 0.097}$ & 0.904 \\
    \midrule
        \multicolumn{5}{l}{\textbf{LLM (BM25+Semantic-CE)}} \\
    \midrule

        Qwen2.5:14b-instruct              & 0.897 & 0.816 & $\boldsymbol{0.843\pm 0.059}$ & 0.894 \\

    Qwen2.5:32b             & 0.850 & 0.933 & $\boldsymbol{0.884\pm 0.058}$ & 0.913 \\
    \midrule
\multicolumn{5}{l}{\textbf{IE + deterministic Rules}} \\
    \midrule
    --             & 0.755 & 0.882 & $\boldsymbol{0.807\pm 0.056}$ & 0.848 \\
    \bottomrule
  \end{tabular*}
\end{table}

\subsubsection{Influence of Predicate Extraction.}
A comparison between CISC-oriented and MCSR-oriented approaches shows that a significant performance gain is not primarily attributable to the logical backend. Rather, the choice of predicate recognition has a greater influence.
Although these variants use the same logic,
the F$_1$ value for the MCSR-oriented approach increases by around nine percentage points compared to the CISC-oriented approach.

\subsubsection{Influence of Fuzzy Logic in LTN Decision-Making.}
The evaluation of the various fuzzy logics reveals a differentiated
picture: For MCSR-BestConf, the Product norm achieves the best results in our data set,
while for CISC and MCSR-TopProb, {\L}ukasiewicz logic has a slight advantage.

\subsubsection{Fault Pattern and Robustness.}
A manual review of incorrectly recognised offers mainly reveals borderline cases with atypical layouts. False positives predominantly originate from invoice-like documents with headers that resemble offers or responses to offers dealing with this topic. The error patterns remain stable across all folds. This indicates a robust combination of LLM-based perception and offer rules.}

\section{Conclusion\opt{long}{ and Outlook}}
\label{sec:conclusion}

In this article, we have presented a neurosymbolic pipeline for validating offers in a regulated context. This pipeline enables text evidence to be linked to logical predefined rules, making the decision taken explainable.
Our experiments with a real corpus of potential offers in the described context show that the performance of the proposed pipeline is comparable to the described baseline models. The strength of the proposed pipeline lies in its interpretability and modular predicate extraction. In practice, the pipeline described above can now be used as a building block to design automation workflows that either reject applications containing invalid offers outright or flag them for review. This can reduce the workload on case handlers.

However, the approach remains limited by several factors: The data set is relatively small, annotated by a single person and comes from a specific institutional and legal context, the rules are domain-specific and hand-coded, and the LLM-based predicate estimators are currently only tested for the German language. Therefore, we understand the results as proof of concept for explainable offer validation in regulated contexts.

\opt{long}{Future work should extend the method to larger and more diverse data sets and expand the internal logic and queries to other domains. In the long run, it seems promising to use neurosymbolic architectures such as the pipeline presented here to make automated decisions in public administration systematically more transparent and verifiable. No out-of-domain (OOD) evaluation has been carried out. This should also be the subject of future work.}

\subsubsection*{Statement of the Authors.}
GPT-5.1 (OpenAI) was used to assist with wording, stylistic revision, code support, prompt design refinement used in the experiments (LLM queries) and the creation of preliminary summaries of external literature. A translation programme (DeepL) was used to assist with the translation from German into English for this text.

\bibliographystyle{splncs04}
\opt{short}{\clearpage}
\bibliography{references}

\begin{thebibliography}{10}
\providecommand{\url}[1]{\texttt{#1}}
\providecommand{\urlprefix}{URL }
\providecommand{\doi}[1]{https://doi.org/#1}

\bibitem{Badreddine.2022}
Badreddine, S., {d'Avila Garcez}, A., Serafini, L., Spranger, M.: Logic tensor
  networks. Artificial Intelligence  \textbf{303},  103649 (2022).
  \doi{10.1016/j.artint.2021.103649}

\bibitem{Bai.2023}
Bai, J., Bai, S., Chu, Y., Cui, Z., Dang, K., Deng, X., Fan, Y., Ge, W., Han,
  Y., Huang, F., Hui, B., Ji, L., Li, M., Lin, J., Lin, R., Liu, D., Liu, G.,
  Lu, C., Lu, K., Ma, J., Men, R., Ren, X., Ren, X., Tan, C., Tan, S., Tu, J.,
  Wang, P., Wang, S., Wang, W., Wu, S., Xu, B., Xu, J., Yang, Yang, H., Yang,
  J., Yang, S., Yao, Y., Yu, B., Yuan, H., Yuan, Z., Zhang, J., Zhang, X.,
  Zhang, Y., Zhang, Z., Zhou, C., Zhou, J., Zhou, X., Zhu, T.: Qwen technical
  report (2023), \url{https://arxiv.org/pdf/2309.16609}

\bibitem{Besold.2022}
Besold, T.R., {d'Avila Garcez}, A., Bader, S., Bowman, H., Domingos, P.,
  Hitzler, P., K{\"u}hnberger, K.U., Lamb, L.C., Lima, P.M.V., de~Penning, L.,
  Pinkas, G., Poon, H., Zaverucha, G.: Neural-symbolic learning and reasoning:
  A survey and interpretation. In: Hitzler, P., Sarker, M.K. (eds.)
  Neuro-symbolic artificial intelligence: the state of the art, chap.~1.
  Frontiers in Artificial Intelligence and Applications, IOS Press, Amsterdam
  and Berlin and Washington, DC (2022). \doi{10.3233/FAIA210348}

\bibitem{Bodhwani.2025}
Bodhwani, U., Ling, Y., Dong, S., Feng, Y., Li, H., Goyal, A.: A calibrated
  reflection approach for enhancing confidence estimation in {LLM}s. In: Cao,
  T., Das, A., Kumarage, T., Wan, Y., Krishna, S., Mehrabi, N., Dhamala, J.,
  Ramakrishna, A., Galystan, A., Kumar, A., Gupta, R., Chang, K.W. (eds.)
  Proceedings of the 5th Workshop on Trustworthy NLP (TrustNLP 2025). pp.
  399--411. {Association for Computational Linguistics}, Stroudsburg, PA, USA
  (2025). \doi{10.18653/v1/2025.trustnlp-main.26}

\bibitem{Devlin.2019}
Devlin, J., Chang, M.W., Lee, K., Toutanova, K.: Bert: Pre-training of deep
  bidirectional transformers for language understanding. In: Burstein, J.,
  Doran, C., Solorio, T. (eds.) Proceedings of the 2019 Conference of the North
  American Chapter of the Association for Computational Linguistics: Human
  Language Technologies, Volume 1 (Long and Short Papers). pp. 4171--4186.
  {Association for Computational Linguistics}, Minneapolis, Minnesota (2019).
  \doi{10.18653/v1/N19-1423}, \url{https://aclanthology.org/N19-1423/}

\bibitem{Donadello.2017}
Donadello, I., Serafini, L., {d'Avila Garcez}, A.: Logic tensor networks for
  semantic image interpretation. In: Sierra, C. (ed.) International Joint
  Conferences on Artificial Intelligence (IJCAI 2017). pp. 1596--1602. {Curran
  Associates Inc}, Red Hook, NY (2017). \doi{10.24963/ijcai.2017/221}

\bibitem{Garcez.2023}
Garcez, A.d., Lamb, L.C.: Neurosymbolic {AI}: the 3rd wave. Artificial
  Intelligence Review  \textbf{56}(11),  12387--12406 (2023).
  \doi{10.1007/s10462-023-10448-w}

\bibitem{Hajek.1998}
H{\'a}jek, P.: Metamathematics of fuzzy logic, Trends in logic, vol.~4. {Kluwer
  Academic}, Dordrecht (1998). \doi{10.1007/978-94-011-5300-3}

\bibitem{huang2022layoutlmv3}
Huang, Y., Lv, T., Cui, L., Lu, Y., Wei, F.: Layoutlmv3: Pre-training for
  document ai with unified text and image masking. In: Proceedings of the 30th
  ACM International Conference on Multimedia. pp. 4083--4091. MM '22,
  Association for Computing Machinery (2022). \doi{10.1145/3503161.3548112}

\bibitem{kim2022donut}
Kim, G., Hong, T., Yim, M., Nam, J., Park, J., Yim, J., Hwang, W., Yun, S.,
  Han, D., Park, S.: Ocr-free document understanding transformer. In: Computer
  Vision -- ECCV 2022. Lecture Notes in Computer Science, vol. 13688, pp.
  498--517. Springer (2022). \doi{10.1007/978-3-031-19815-1_29}

\bibitem{Kohavi.1995}
Kohavi, R.: A study of cross-validation and bootstrap for accuracy estimation
  and model selection. In: Proceedings of the 14th International Joint
  Conference on Artificial Intelligence -- Volume 2. pp. 1137--1143. IJCAI'95,
  {Morgan Kaufmann Publishers Inc}, San Francisco, CA, USA (1995).
  \doi{10.5555/1643031.1643047}

\bibitem{Lewis.2020}
Lewis, P., Perez, E., Piktus, A., Petroni, F., Karpukhin, V., Goyal, N.,
  K{\"u}ttler, H., Lewis, M., Yih, W.t., Rockt{\"a}schel, T., Riedel, S.,
  Kiela, D.: Retrieval-augmented generation for knowledge-intensive {NLP}
  tasks. In: {H. Larochelle}, {M. Ranzato}, {R. Hadsell}, {M.F. Balcan}, {H.
  Lin} (eds.) Advances in Neural Information Processing Systems. vol.~33, pp.
  9459--9474. {Curran Associates, Inc} (2020),
  \url{https://dl.acm.org/doi/proceedings/10.5555/3495724}

\bibitem{Lin.2022}
Lin, J., Nogueira, R., Yates, A.: Pretrained Transformers for Text Ranking:
  BERT and Beyond. Synthesis Lectures on Human Language Technologies, {Springer
  International Publishing} and {Imprint Springer}, Cham, 1st ed. 2022 edn.
  (2022). \doi{10.1007/978-3-031-02181-7}

\bibitem{manhaeve2021deepproblog}
Manhaeve, R., Duman{\v{c}}i{\'c}, S., Kimmig, A., Demeester, T., De~Raedt, L.:
  Neural probabilistic logic programming in {DeepProbLog}. Artificial
  Intelligence  \textbf{298},  103504 (2021).
  \doi{10.1016/j.artint.2021.103504}

\bibitem{Powers.2011}
Powers, D.M.W.: Evaluation: from precision, recall and {F}-measure to {ROC},
  informedness, markedness and correlation. International Journal of Machine
  Learning Technology  (2011), \url{https://arxiv.org/pdf/2010.16061}

\bibitem{Richmond.2023}
Richmond, K.M., Muddamsetty, S.M., Gammeltoft-Hansen, T., Olsen, H.P.,
  Moeslund, T.B.: Explainable {AI} and law: An evidential survey. Digital
  Society  \textbf{3}(1), ~1 (2023). \doi{10.1007/s44206-023-00081-z}

\bibitem{Robertson.2009}
Robertson, S., Zaragoza, H.: The probabilistic relevance framework: {BM25} and
  beyond. Foundations and Trends$^\text{\textregistered}$ in Information
  Retrieval  \textbf{3}(4),  333--389 (2009). \doi{10.1561/1500000019}

\bibitem{Rudin.2019}
Rudin, C.: Stop explaining black box machine learning models for high stakes
  decisions and use interpretable models instead. Nature Machine Intelligence
  \textbf{1}(5),  206--215 (2019). \doi{10.1038/s42256-019-0048-x}

\bibitem{Serafini.2016}
Serafini, L., {d'Avila Garcez}, A.S.: Learning and reasoning with logic tensor
  networks. In: Adorni, G., Cagnoni, S., Gori, M., Maratea, M. (eds.) AI*IA
  2016: advances in artificial intelligence, Lecture Notes in Artificial
  Intelligence, vol. 10037, pp. 334--348. Springer, Cham (2016).
  \doi{10.1007/978-3-319-49130-1_25}

\bibitem{Sokolova.2009}
Sokolova, M., Lapalme, G.: A systematic analysis of performance measures for
  classification tasks. Information Processing {\&} Management  \textbf{45}(4),
   427--437 (2009). \doi{10.1016/j.ipm.2009.03.002}

\bibitem{Taubenfeld.2025}
Taubenfeld, A., Sheffer, T., Ofek, E., Feder, A., Goldstein, A., Gekhman, Z.,
  Yona, G.: Confidence improves self-consistency in {LLM}s. In: Che, W.,
  Nabende, J., Shutova, E., Pilehvar, M.T. (eds.) Findings of the Association
  for Computational Linguistics: ACL 2025. pp. 20090--20111. {Association for
  Computational Linguistics}, Stroudsburg, PA, USA (2025).
  \doi{10.18653/v1/2025.findings-acl.1030}

\bibitem{yang2020neurasp}
Yang, Z., Ishay, A., Lee, J.: Neurasp: Embracing neural networks into answer
  set programming. In: Proceedings of the Twenty-Ninth International Joint
  Conference on Artificial Intelligence. pp. 1755--1762 (2020).
  \doi{10.24963/ijcai.2020/243}

\end{thebibliography}

\opt{long}{\appendix
\section*{Appendix}
\fontsize{8.5}{10}\selectfont

\section{Information Extraction (IE) Pattern Groups}
\label{tab:ie-patterns}
\centering
\begin{tabular}{p{0.18\linewidth}p{0.74\linewidth}}
\toprule
\textbf{Score} & \textbf{Pattern group} \\
\midrule
$s_T$ &
\texttt{angebot}, \texttt{offerte}, \texttt{quotation}, \texttt{offer},
\texttt{quote}, \texttt{kostenvor\-anschlag}, \texttt{preisinformation}.
\\
\midrule
$s_N$ &
\texttt{angebot(s)nummer}, \texttt{angebot(s)-nr.}, \texttt{quotation no},
\texttt{quotation nr}, \texttt{quotation number}, \texttt{quote id},
\texttt{quote no}, \texttt{quote nr}, \texttt{quote number},
\texttt{offer id}, \texttt{offer no}, \texttt{offer nr},
\texttt{offer number}, \texttt{referenz}, \texttt{reference}
\\
\midrule
$s_V$ &
\texttt{gültig}, \texttt{gueltig}, \texttt{gültigkeit},
\texttt{gueltigkeit}, \texttt{valid}, \texttt{validity},
\texttt{valid until}, \texttt{gültig bis}, \texttt{gueltig bis},
\texttt{angebot gilt}; date patterns
\texttt{dd.mm.yyyy}, \texttt{dd-mm-yyyy}, \texttt{dd/mm/yyyy},
and month-name dates.
\\
\midrule
$s_R$ &
\texttt{freibleibend}, \texttt{unverbindlich},
\texttt{subject to availability}, \texttt{subject to prior sale},
\texttt{zwischenverkauf vorbehalten}, \texttt{non-binding},
\texttt{ohne gewähr}.
\\
\midrule
$s_P$ &
\texttt{zahlungsbedingungen}, \texttt{payment terms},
\texttt{zahlbar}, \texttt{netto}, \texttt{skonto},
\texttt{prepayment}, \texttt{vorkasse}, \texttt{due within},
\texttt{zahlung innerhalb}; concrete-term patterns:
\texttt{tage}, \texttt{days}, \texttt{netto}, \texttt{skonto},
\texttt{within}.
\\
\midrule
$s_D$ &
\texttt{lieferbedingungen}, \texttt{lieferzeit}, \texttt{lieferung},
\texttt{delivery terms}, \texttt{delivery time}, \texttt{shipping terms},
\texttt{incoterms}, \texttt{exw}, \texttt{fob}, \texttt{cif},
\texttt{cip}, \texttt{dap}, \texttt{ddp}, \texttt{dpu},
\texttt{fca}, \texttt{cfr}; concrete-term patterns:
\texttt{tage}, \texttt{days}, Incoterms.
\\
\midrule
$s_S$ &
\texttt{ansprechpartner}, \texttt{kontakt}, \texttt{vertrieb},
\texttt{sales}, \texttt{account manager}, \texttt{your contact},
\texttt{ihr kontakt}; e-mail pattern
\texttt{[\textbackslash w.+-]+@[\textbackslash w.-]+\textbackslash.[a-z]\{2,\}}.
\\
\midrule
$s_{\neg O}$ &
\texttt{rechnung}, \texttt{invoice}, \texttt{lieferschein},
\texttt{delivery note}, \texttt{bestellung}, \texttt{purchase order},
\texttt{auftragsbestätigung}, \texttt{order confirmation},
\texttt{warenkorb}, \texttt{shopping cart}, \texttt{cart},
\texttt{proforma invoice}, \texttt{gutschrift}, \texttt{credit note}.
\\
\bottomrule
\end{tabular}

\section{Document Preprocessing and Retrieval Configuration}
\label{tab:repro-preprocessing-retrieval}
\centering
\renewcommand{\arraystretch}{1.12}
\begin{tabular}{@{}p{0.36\linewidth}p{0.56\linewidth}@{}}
\toprule
\textbf{Name} & \textbf{Value} \\
\midrule

OCR engine &
Tesseract OCR \\

OCR languages &
\texttt{deu+eng} \\

OCR resolution &
220 DPI \\

Maximum OCR pages &
30 pages per document \\

Embedding model &
\texttt{nomic-embed-text} \\

Chunking method &
Sentence-based chunking \\

Chunk size &
450-token budget \\

Chunk overlap &
1 sentence \\

BM25 candidate cutoff &
$K_1 = 200$ \\

Embedding reranking cutoff &
$K_2 = 40$ \\

Final predicate chunks &
$K = 6$ chunks per predicate \\

\bottomrule
\end{tabular}}

\end{document}